\documentclass[runningheads]{llncs}

\usepackage[T1]{fontenc}
\usepackage{graphicx}
\usepackage[colorlinks=true, linkcolor=black, citecolor=black, urlcolor=black]{hyperref}

\usepackage{amsmath}
\usepackage{array}
\usepackage{float}
\usepackage{graphicx}
\usepackage{longtable}
\usepackage{multirow}
\usepackage{subcaption}
\usepackage{caption}

\begin{document}
\title{KGRAG-Ex: Explainable Retrieval-Augmented Generation with Knowledge Graph-based Perturbations}
\titlerunning{KGRAG-Ex: Explainable RAG with KG-based Perturbations}
%
\author{Georgios Balanos\inst{1,2} \and
Evangelos Chasanis\inst{1,2} \and
Konstantinos Skianis\inst{1} \and
Evaggelia Pitoura\inst{1,2}
}
\authorrunning{G. Balanos et al.}
%
\institute{University of Ioannina, Greece \and
Archimedes, Athena Research Center, Greece
}
\maketitle              
\begin{abstract}
Retrieval-Augmented Generation (RAG) enhances language models by grounding responses in external information, yet explainability remains a critical challenge—particularly when retrieval relies on unstructured text. Knowledge graphs (KGs) offer a solution by introducing structured, semantically rich representations of entities and their relationships, enabling transparent retrieval paths and interpretable reasoning. In this work, we present KGRAG-Ex, a RAG system that improves both factual grounding and explainability by leveraging a domain-specific KG constructed via prompt-based information extraction. Given a user query, KGRAG-Ex identifies relevant entities and semantic paths in the graph, which are then transformed into pseudo-paragraphs—natural language representations of graph substructures that guide corpus retrieval. To improve interpretability and support reasoning transparency, we incorporate perturbation-based explanation methods that assess the influence of specific KG-derived components on the generated answers. We conduct a series of experiments to analyze the sensitivity of the system to different perturbation methods, the relationship between graph component importance and their structural positions, the influence of semantic node types, and how graph metrics correspond to the influence of components within the explanations process.


\end{abstract}
\section{Introduction}
Large Language Models (LLMs) \cite{naveed2024comprehensiveoverviewlargelanguage} have demonstrated strong performance across numerous natural language processing tasks, including question answering (QA) \cite{kamalloo-etal-2023-evaluating}, summarization \cite{zhangTACLsummarization}, and dialogue generation \cite{zhao-etal-2020-knowledge-grounded}. While LLMs excel at generating fluent and contextually relevant text, their reliance on internalized knowledge can lead to factual errors and untraceable reasoning. To mitigate this, Retrieval-Augmented Generation (RAG) \cite{lewisNEURIPS2020} combines LLMs with external document retrieval, grounding responses in up-to-date information drawn from large corpora. However, RAG systems typically offer limited visibility into how retrieved content influences the output of the model, raising concerns in settings where outputs must be auditable, interpretable, and verifiable.

Explainability in RAG has thus become a key focus of recent research. Approaches such as \cite{10.1145/3626772.3657660} and \cite{10598017} use perturbation-based techniques to analyze how altering parts of the input—such as retrieved documents or query terms—affects the final output. These methods provide insights into the importance of specific elements in the context, helping to trace the decision-making process of the model. However, they primarily operate on unstructured text and offer limited access to explicit reasoning pathways or underlying semantic structures. Additionally, these techniques can be computationally expensive, which limits their scalability and practicality in real-world deployments.

In this work, we propose KGRAG-Ex, a novel system that enhances explainability in RAG by integrating structured knowledge graphs (KGs) \cite{zhu2025knowledge} into the retrieval and generation pipeline \cite{bockling2024walkretrieve}. The system processes user queries to identify relevant entities, constructs semantic paths through the KG, and converts these paths into natural language pseudo-paragraphs that guide document retrieval. These structured representations provide a transparent layer of reasoning that complements the unstructured evidence retrieved from external sources.

A key novelty of our approach lies in combining structured retrieval via knowledge graph paths with targeted perturbations at the path level to analyze and explain model outputs. Unlike previous perturbation-based methods that primarily operate on unstructured text, our method leverages explicit reasoning pathways embedded in knowledge graphs to provide more transparent, faithful, and fine-grained explanations. This integration enhances interpretability while also improving computational efficiency and scalability, making it especially suitable for applications requiring auditable and trustworthy reasoning.

We perform perturbations by removing nodes, edges, or  sub-paths.
By modifying nodes, edges, or sub-paths, we evaluate how these structural changes influence the output of the model. This KG-focused perturbation approach provides a more comprehensive and fine-grained view of model behavior, highlighting which elements of the graph-based reasoning process are most critical. Furthermore, by isolating the impact of specific graph components, KGRAG-Ex enhances interpretability  without relying on text-level perturbations, as in previous work. Perturbations at the path level tend also to be more cost-efficient than perturnations at the text level, making the framework particularly suitable for high-stakes domains where reasoning traceability is essential. 

Finally, we conduct experiments to analyze the sensitivity to different perturbation methods, the influence of semantic node types, and the relationship between graph component importance,   their structural positions in the paths and centrality measures in the KG. We also compare the efficiency of our approach to text-based perturbation methods.


The remainder of this paper is structured as follows. Section~\ref{sec:related-work} reviews related work. 
Section~\ref{sec:kg-construction} describes our method for knowledge graph construction and retrieval, and
Section~\ref{sec:explanations} presents our perturbation-based explanation framework.
Section~\ref{sec:experiments} reports experimental results, and  Section~\ref{sec:conclusion} concludes the paper.


\section{Related Work}
\label{sec:related-work}

Large language models (LLMs) have been used to extract structured representations—such as entity–relation–entity triples—from unstructured text, enabling the construction of dynamic knowledge graphs to guide retrieval and reasoning \cite{zhu2025knowledge}. 
Query2doc \cite{query2doc} demonstrates how LLMs can be prompted to expand sparse queries into detailed pseudo-documents. In this paper, rather than a generic expansion, we extend this approach by generating structured pseudo-paragraphs based on the extracted entities and their connected paths within the KG. These paragraphs are then combined into a pseudo-document that serves as a context-rich input to guide retrieval.

Recent advancements in explaining Retrieval-Augmented Generation (RAG) models have been addressed by works such as \cite{10.1145/3626772.3657660} and \cite{10598017}. RAG-Ex  \cite{10.1145/3626772.3657660} introduces a model- and language-agnostic framework that provides approximate explanations for why large language models (LLMs) generate specific responses in RAG-based question answering tasks. By employing perturbation-based methods, RAG-Ex identifies critical tokens whose removal significantly alters the output, offering insights into the decision-making process of LLMs. In contrast, RAGE \cite{10598017} focuses on the provenance of external knowledge sources used during RAG. It employs counterfactual explanations by systematically removing or reordering retrieved documents to observe changes in the model's responses, thereby highlighting the influence of specific sources on the generated answers. 

Our work is situated within a broader landscape of explainability in NLP. Surveys such as \cite{danilevsky2020survey} provide a comprehensive taxonomy of explanation methods in NLP, while \cite{alkhamissi2022review} reviews interpretability techniques for large language models (LLMs), including prompt-based and retrieval-augmented models. More recently, \cite{belinkov2022probing} explores how probing and intervention methods can be applied to understand the internal mechanisms of LLMs, offering complementary perspectives to perturbation-based explanations. Building on prior work such as RAG-Ex and RAGE, which apply perturbation-based techniques to unstructured retrieval and document provenance, respectively, our approach—KGRAG-Ex—introduces structured, graph-based retrieval via knowledge graphs. This enables more interpretable reasoning paths and targeted perturbation of graph-derived information, leading to more focused and transparent explanations.


\section{Knowledge Graph Construction and Retrieval}
\label{sec:kg-construction}

To support structured retrieval and enhance reasoning transparency, we construct a domain-specific knowledge graph - focusing on the medical domain in this paper - directly from raw textual data.
Given a user query, we extract relevant paths from the knowledge graph and convert them into contextual input for RAG.

\paragraph{Knowledge-Graph Construction.}

\begin{figure}
    \centering
    \includegraphics[width=1\linewidth]{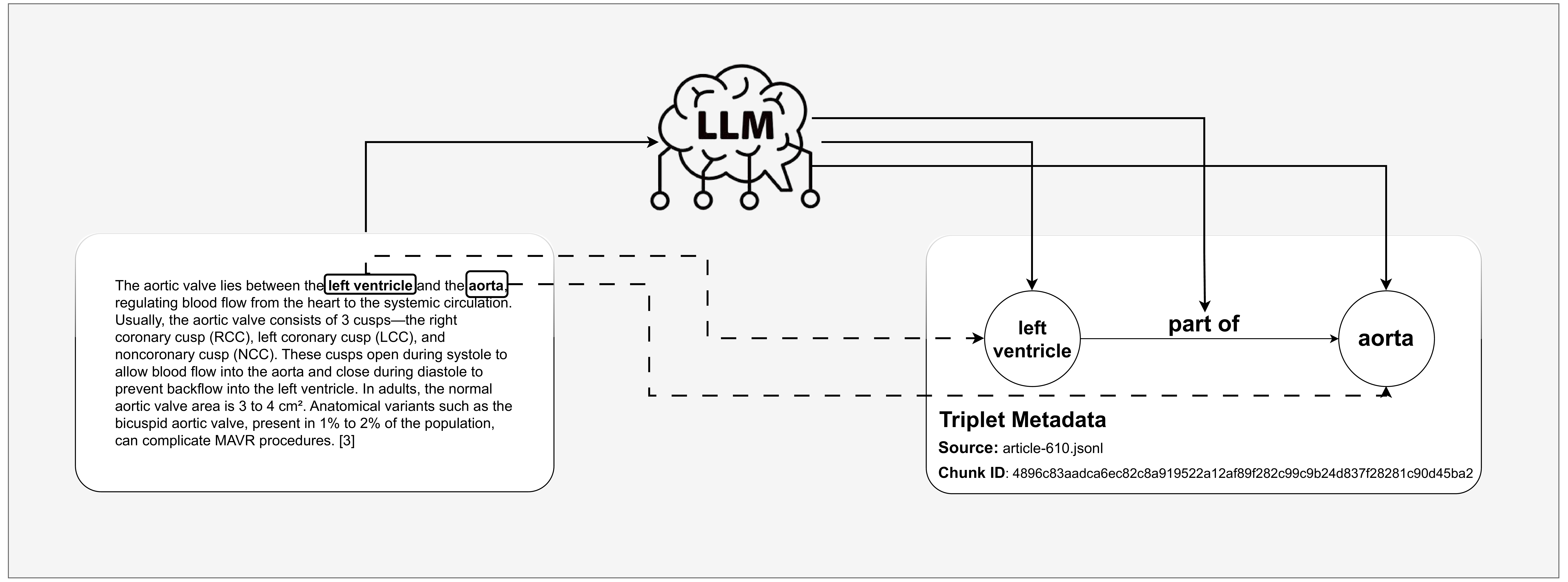}
    \caption{Triplet extraction.}
    \label{fig:triplet-extraction}
\end{figure}

To construct the knowledge graph, consistent with existing studies in retrieval-augmented generation (RAG) (Lewis et al., 2020~\cite{DBLP:journals/corr/abs-2005-11401}; Gao et al., 2023~\cite{jiang2023activeretrievalaugmentedgeneration}; Fan et al., 2024~\cite{fan2024surveyragmeetingllms}), each document is divided into smaller chunks, respecting a predefined chunk size. For each chunk, we prompt a large language model (LLM), to extract factual information in the form of (Entity, Relationship, Entity) triplets as in \cite{zhu2025knowledge}. Additionally, the LLM assigns a semantic label to each extracted entity (node), such as Disease, Body Part, or Medication, to enrich the graph with meaningful categories.
Each extracted triplet is annotated with metadata, including the document identifier and the chunk index, enabling traceability and contextual alignment within the source corpus. The construction is illustrated in Figure~\ref{fig:triplet-extraction}.

\paragraph{Retrieval from the Knowledge Graph.}

To retrieve relevant information from the knowledge graph, we first extract a set of (Entity, Relationship, Entity) triplets from the user query using an LLM. This step ensures that the query is transformed into a structured form that aligns with the format of the knowledge graph. Entities mentioned in the query are identified and grounded to their corresponding nodes in the graph. Once the relevant entities are identified, we perform graph traversal to determine the shortest paths between them within the knowledge graph. This process uncovers the intermediate nodes and relations that connect the query-relevant entities, allowing us to identify semantically meaningful subgraphs. If no shortest path is found between the extracted entities, a fallback mechanism is implemented where regular retrieval methods (without using the knowledge graph) are applied to ensure robust information access. 
The resulting graph structure includes not only the semantic connections between entities but also metadata such as the source filename and chunk ID for each edge. This additional information enhances traceability, allowing each element of the graph to be linked back to its original location in the source documents. 

An example of the retrieval process, applied to our running example of a medical dataset with multiple-choice  QA, is presented in Table \ref{tab:first-example-perturbations}, which displays first the user query and the available options. For this specific query, we identified the shortest path from \textit{pulmonary hypoplasia} to \textit{oligohydramnios}. Each triplet in the derived path includes the source file names, thereby enhancing the traceability of our system.

\begin{table}[ht]
\caption{A retrieval example: the initial user-query, the selected option,  the retrieved path, and and related sources.}
\label{tab:first-example-perturbations}
\centering
\footnotesize
\begin{tabular}{>{\raggedright\arraybackslash}p{4cm}||p{8cm}}
\hline\hline
\textbf{Query} & A baby born with pulmonary hypoplasia secondary to oligohydramnios caused by renal agenesis would be classified as having\newline
\textbf{Options:} \newline
A. an association \quad B. a dysplasia\newline
C. a sequence \quad D. a syndrome \\
\hline
\textbf{Selected Option:} & A. an association \\
\hline\hline
\textbf{Retrieved Path} &
- pulmonary hypoplasia \(\xrightarrow{\text{IS RISK FACTOR FOR}}\) persistent pulmonary hypertension in the newborn \\[0.5em]
& - persistent pulmonary hypertension in the newborn \(\xrightarrow{\text{HAS RISK FACTOR}}\) oligohydramnios \\
\hline\hline
\textbf{Sources:} & \textbf{Article source: article-27634.jsonl} \newline
\textbf{Article source: article-22355.jsonl} \\
\hline\hline
\end{tabular}
\end{table}


\paragraph{Transforming Knowledge Graph Paths into RAG Context.}

We construct a context as a means of translating structured data into a form amenable to language-based models. The process begins with the generation of pseudo-paragraphs from structured paths —such as sequences of nodes, concepts, or semantic transitions. Each pseudo-paragraph captures the meaning of its corresponding path in natural language, serving as a textual proxy for the underlying structure.

Once a collection of pseudo-paragraphs has been generated, these are concatenated to form a single context. This context integrates information from multiple paths, providing a unified textual representation that reflects the overall semantics of the structured source. By converting structured inputs into coherent natural language, this method enables the effective use of large language models and other NLP systems for tasks involving structured or semi-structured data. The resulting textual context is then used as input for retrieval-augmented generation (RAG), providing relevant and comprehensive information to improve the performance of downstream language tasks. 

\begin{figure}
    \centering
    \includegraphics[width=1\linewidth]{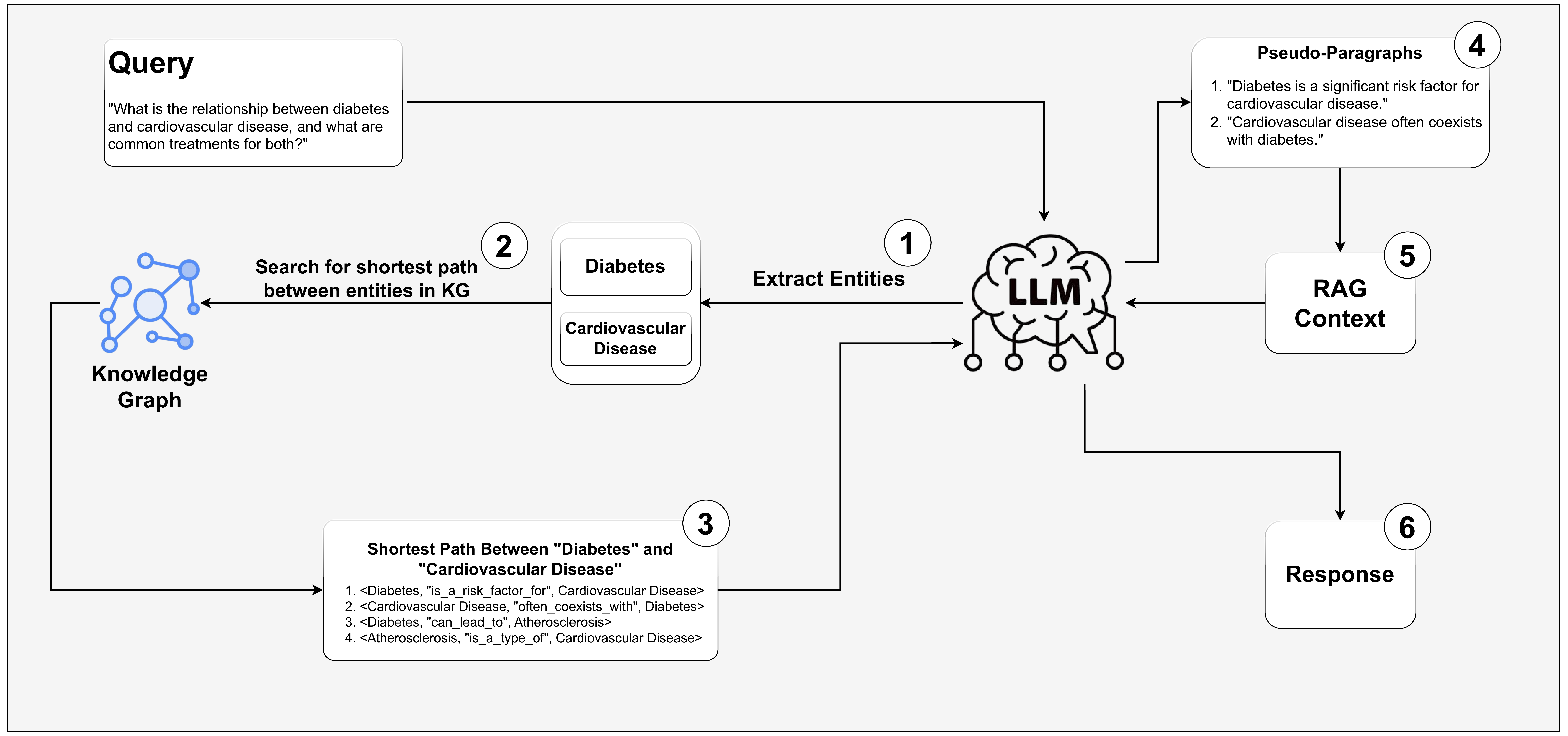}
    \caption{The proposed pipeline. Graph-level perturbations are applied at Step 3.}
    \label{fig:execution-example}
\end{figure}

Figure \ref{fig:execution-example} illustrates our proposed pipeline. This includes entity extraction from the user query, computation of the derived shortest paths, generation of pseudo-paragraphs, and augmentation of the RAG context.

\captionsetup{labelfont=bf}

\section{Explanations through Knowledge Graph Perturbations}
\label{sec:explanations}

To generate explanations, we apply perturbations to the retrieved knowledge graph context by selectively removing elements at varying levels of granularity, this includes individual nodes, edges, or entire sub-paths. The motivation behind this approach is to identify which components of the knowledge graph are most influential in the reasoning process. By systematically deleting specific elements and observing the resulting changes in model predictions or outputs, we can infer their importance: if removing a component significantly alters the outcome, it suggests that the component plays a critical role. These controlled perturbations enable us to assess the semantic contribution of each part of the reasoning chain, ultimately facilitating the generation of faithful and interpretable explanations. Since every graph component is—directly or indirectly—annotated with metadata during knowledge graph construction, these perturbation methods also enhance the traceability of the model by linking important elements—nodes, edges, or sub-paths—back to their sources.

\paragraph{Node-Level Perturbations.}
In this perturbation method, we remove one node from the derived path while keeping all other elements unchanged. By comparing the output of the model  with and without the node, we can estimate the contribution of each entity to the final response. Furthermore, since every node is labeled with a semantic category during the knowledge graph construction (e.g. Disease, Medication, Body Parts), this analysis reveals not only which individual nodes are influential, but also which types of information the model relies on the most. This may offer insights into the semantic reasoning patterns of the model and help identify the categories of knowledge that drive its outputs.

\paragraph{Edge-Level Perturbations.}
In this method, each perturbation corresponds to the removal of a single edge from the derived path, disconnecting two semantically related nodes. While node-level perturbations highlight important entities, edge-level perturbations reveal how the model fills the semantic space between them, that is, how it connects these entities through meaningful relationships. Since edges often represent actions or associations (e.g., IS DETECTED BY, IS AFFECTED BY), their removal helps identify which specific relations are most critical to  reasoning. In many cases, this points to the key interaction or action that drives the response to a given query.

\paragraph{Sub-path-Level Perturbations.}
For sub-path-level perturbations, we remove a sub-path, typically a triplet, from the derived path. Compared to node- and edge-level perturbations, these changes tend to have a greater impact on the reasoning path, as they involve the removal of multiple elements at once.

\begingroup
\footnotesize
\begin{longtable}{>{\raggedright\arraybackslash}p{4cm}||p{8cm}}
\caption{Example perturbations on the retrieved path shown in Table \ref{tab:first-example-perturbations}.} \label{tab:example-perturbations} \\
\hline\hline
\textbf{Path after Node Perturbation} & \textbf{Removed Node: persistent pulmonary hypertension in the newborn} \\
& - pulmonary hypoplasia \(\xrightarrow{\text{IS RISK FACTOR FOR}}\) \\[0.5em]
& -\(\xrightarrow{\text{HAS RISK FACTOR}}\) oligohydramnios \\
\hline
\textbf{Selected Option} & C. a sequence \\
\hline\hline

\textbf{Path after Edge Perturbation} & \textbf{Removed Edge: HAS RISK FACTOR} \\
& - pulmonary hypoplasia \(\xrightarrow{\text{IS RISK FACTOR FOR}}\) persistent pulmonary hypertension in the newborn \\[0.2em]
& - persistent pulmonary hypertension in the newborn -- oligohydramnios \\
\hline
\textbf{Selected Option} & C. a sequence \\
\hline\hline

\textbf{Path after Subpath Perturbation} & \textbf{Removed Subpath:} \newline 
persistent pulmonary hypertension in the newborn \(\xrightarrow{\text{HAS RISK FACTOR}}\) oligohydramnios \\[1.5em]
& - pulmonary hypoplasia \(\xrightarrow{\text{IS RISK FACTOR FOR}}\) persistent pulmonary hypertension in the newborn \\
\hline
\textbf{Selected Option} & A. an association \\
\hline\hline

\textbf{Article Sources} & article-27634.jsonl \newline article-22355.jsonl \\
\hline\hline

\textbf{Explanation shown to user} & 
The most important condition for answering the question is Persistent Pulmonary Hypertension in the Newborn. It had the biggest impact on the result. \\
\hline\hline

\textbf{System insight (technical)} & 
Removing \textbf{Persistent Pulmonary Hypertension in the Newborn} led to a different answer \textbf{2 times}, indicating it is a highly influential entity in the reasoning path.  
Removing \textbf{Oligohydramnios} led to a different answer \textbf{1 time}, showing it has a moderate impact on the final answer. \\
\hline\hline
\end{longtable}
\endgroup

An example is shown in Table \ref{tab:example-perturbations}. We include one example of each perturbation method and the selected option after the perturbation. We provide two types of explanations, one is targeting end users and the other one is more technical. The explanation shown to the end user is the most influential graph element, identified as the one that causes the most changes in the output of the model across the different perturbations, e.g., the one that appears in most of them. In this example, this is \textit{persistent pulmonary hypertension in the newborn}. 
We also report technical details for developers.
Finally, we obtain the source filename from which \textit{persistent pulmonary hypertension in the newborn} was extracted, further improving traceability by revealing the origin of the most influential context elements.

%
%
%
%
%

\section{Experiments}
\label{sec:experiments}



In our evaluation, we utilized the StatPearls corpus along with two established benchmarks: MedMCQA and MMLU \cite{xiong-etal-2024-benchmarking}. MedMCQA is a multiple--choice question --answering dataset derived from real medical entrance exams, focusing on clinical and factual medical knowledge. MMLU (Massive Multitask Language Understanding) complements this by offering a wide-ranging evaluation across 57 subjects, including medicine, law, mathematics, and more. Both MedMCQA and MMLU consist of multiple-choice questions with four options, enabling consistent evaluation of model reasoning across specialized (medical) and general knowledge domains.

 Due to computational and token-related costs (API token limits) associated with large-scale language model usage, the KG was built from a subset of the full corpus. Table~\ref{tab:kg_overview} summarizes key statistics of the constructed knowledge graph and document corpus, as well as the distribution of entity labels assigned during preprocessing.

The code and data used in our experiments are publicly available\footnote{\url{https://github.com/george-balanos/KGRAG-Ex}}.

\begin{table}[h]
\centering
\captionsetup{labelfont=bf}
\caption{Knowledge Graph and Corpus}
\label{tab:kg_overview}
\begin{minipage}[t]{0.48\textwidth}
\centering
\begin{tabular}{|l|c|}
\hline
\multicolumn{2}{|c|}{\textbf{Document Corpus}} \\
\hline
Total documents           & 9,552 \\
Total text chunks         & 445,454 \\
Average chunks per document & 46.63 \\
Standard deviation        & 28.58 \\
Min chunks per document   & 5 \\
Max chunks per document   & 556 \\
\hline
\multicolumn{2}{|c|}{\textbf{Knowledge Graph}} \\
\hline
Total nodes               & 53,411 \\
Total edges               & 133,287 \\
Average in-degree         & 2.495 \\
Average out-degree        & 2.495 \\
Average total degree      & 4.990 \\
\hline
\end{tabular}
\end{minipage}
\hfill
\begin{minipage}[t]{0.48\textwidth}
\centering
\renewcommand{\arraystretch}{1.455}  
\begin{tabular}{|l|c|}
\hline
\textbf{Entity Label} & \textbf{Percentage (\%)} \\
\hline
Diseases             & 30.73 \\
Risk Factors         & 8.50 \\
Symptoms             & 17.71 \\
Diagnostic Tests     & 9.29 \\
Treatments           & 14.97 \\
Body Parts           & 11.81 \\
Medications          & 6.83 \\
Unknown              & 0.16 \\
\hline
\end{tabular}
\renewcommand{\arraystretch}{1}  
\end{minipage}
\end{table}



\subsection{Evaluation of the Proposed Methodology}

To gain deeper insights into the effectiveness of our graph perturbation strategies, we conduct  experiments to address the following research questions:


\begin{itemize}
    \item[] \textbf{RQ1}: How sensitive is the system to each perturbation method?
    \item[] \textbf{RQ2}: Is there a correlation between perturbed graph elements and their relative positions?
    \item[] \textbf{RQ3}: Which types of nodes, based on their assigned semantic labels, are most critical in the generation process?
    \item[] \textbf{RQ4}: How do graph centrality metrics (e.g., node degree, edge betweenness) relate to the significance of elements within the derived paths?
\end{itemize}

\subsubsection{Evaluating the relative effect of the the graph perturbation methods.}

To assess how different types of graph component removals influence the generated output, we evaluate the sensitivity of the model to different perturbations. Specifically, for each example, we generate and apply multiple perturbations across three types:
removing individual nodes, removing edges, and removing sub-paths consisting of triplets of nodes.

Each perturbation is applied independently, and we observe whether it leads to a change in the generated output. For each example, we compare the frequency of output changes caused by the different perturbation types to determine which type has the greatest influence. By repeating this process across the full evaluation set, we  identify which type of graph modification tends to have the most significant impact. Table~\ref{tab:impactfulness} presents these findings on the MedMCQA and MMLU datasets, showing that sub-path removals generally have the strongest impact, followed by node and edge removals. This observation aligns with our expectations, since sub-path removal tends to induce the most changes within the corresponding path.

\captionsetup{labelfont=bf}
\begin{table}[h]
\centering
\caption{Impact of perturbation methods across datasets}
\label{tab:impactfulness}
\small
\resizebox{0.98\textwidth}{!}{%
\begin{tabular}{|c|c|c|c|c|}
\hline
\textbf{Dataset} & \textbf{Perturbation Examples} & \textbf{Node Impact} & \textbf{Edge Impact} & \textbf{Sub-path Impact} \\
\hline
MedMCQA & 128 & 41 & 30 & 57 \\
MMLU    & 37 & 6 & 8 & 23 \\
\hline
\end{tabular}
}
\end{table}

\subsubsection{Exploring the role of perturbation position along the retrieved path.}

In this experiment, we analyze the relationship between critical perturbations -- i.e., perturbations that lead to output changes -- and their relative position in the shortest path. To enable consistent comparison across paths of different lengths, we normalize the position of each perturbed node, edge, or sub-path on a scale from 0 (start of the path) to 1 (end of the path). This normalization is essential for identifying positional trends independent of absolute path length.

\begin{figure}[h]
    \centering

    \begin{subfigure}[b]{0.32\linewidth}
        \includegraphics[width=\linewidth]{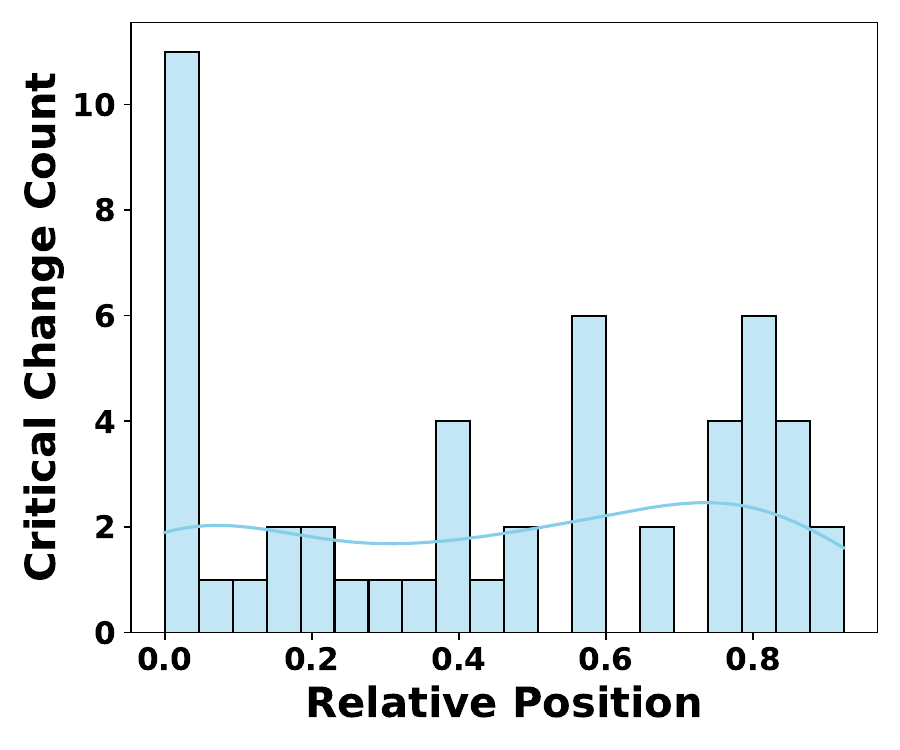}
        \label{fig:node_positions_mmlu}
    \end{subfigure}
    \hfill
    \begin{subfigure}[b]{0.32\linewidth}
        \includegraphics[width=\linewidth]{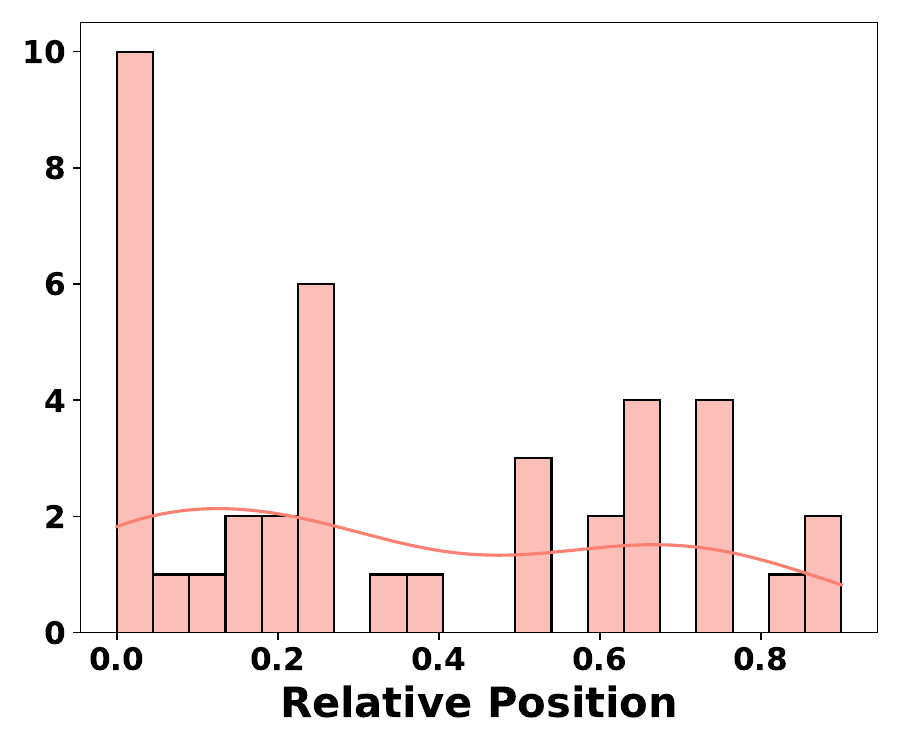}
        \label{fig:edge_positions_mmlu}
    \end{subfigure}
    \hfill
    \begin{subfigure}[b]{0.32\linewidth}
        \includegraphics[width=\linewidth]{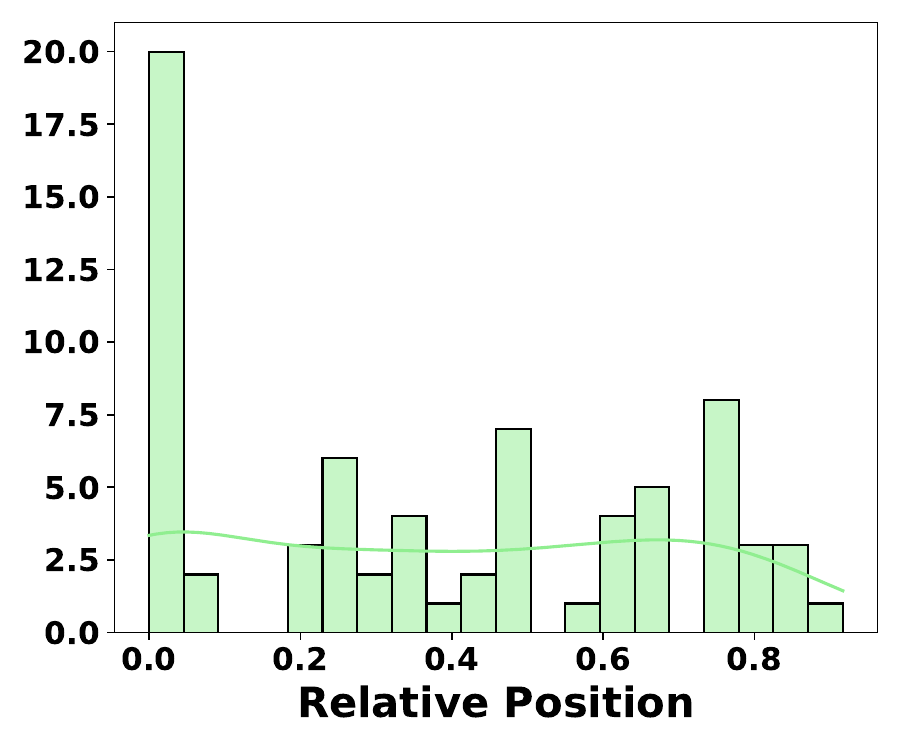}
        \label{fig:subpath_positions_mmlu}
    \end{subfigure}


    \begin{subfigure}[b]{0.32\linewidth}
        \includegraphics[width=\linewidth]{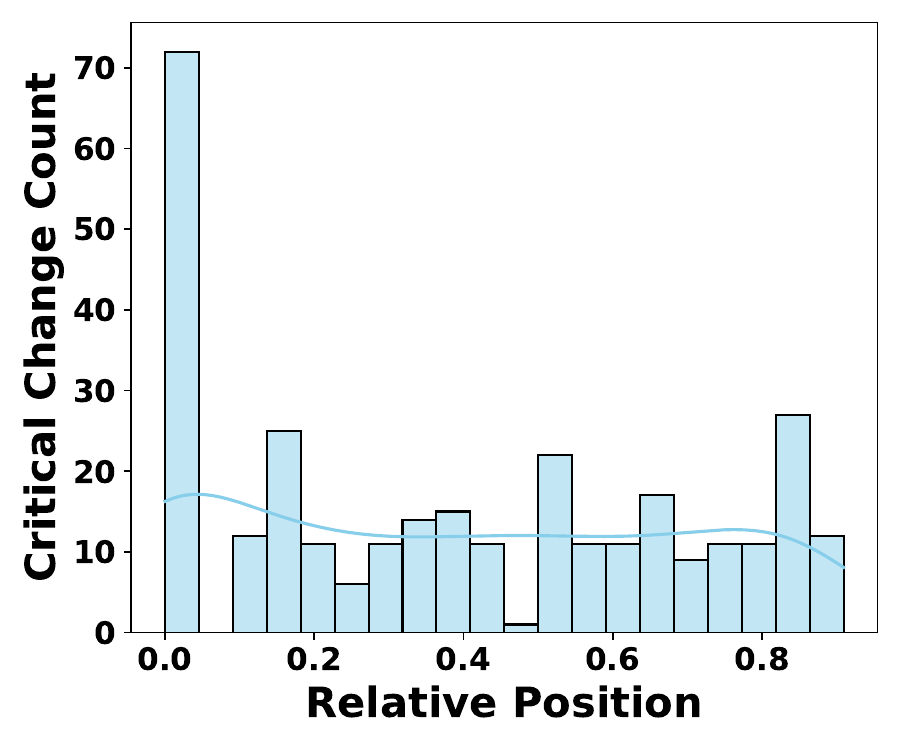}
        \caption{Node Positions}
        \label{fig:node_positions_medmcqa}
    \end{subfigure}
    \hfill
    \begin{subfigure}[b]{0.32\linewidth}
        \includegraphics[width=\linewidth]{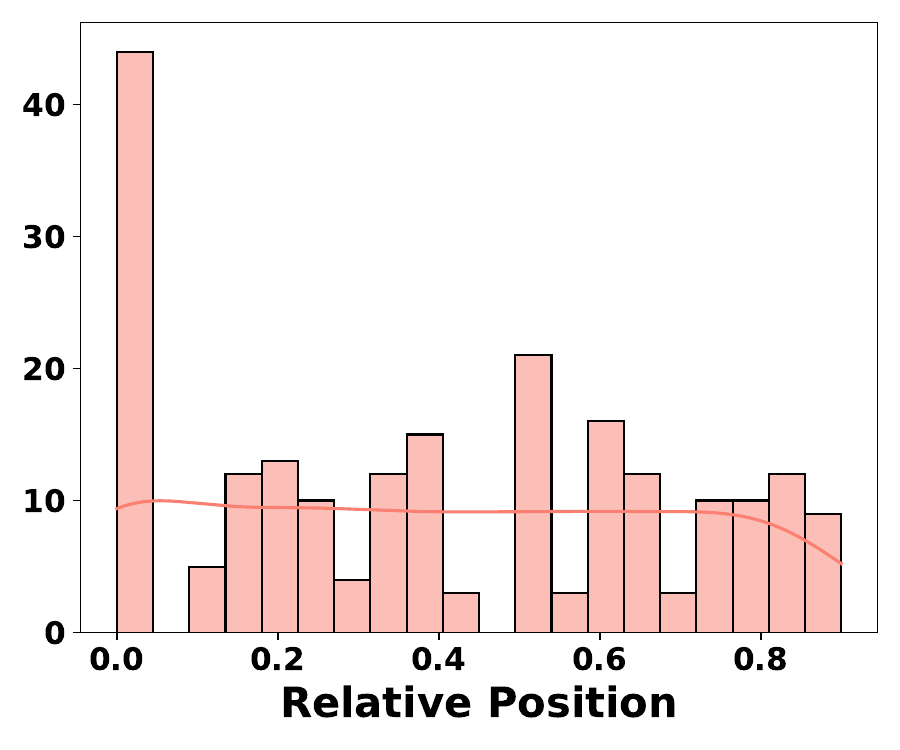}
        \caption{Edge Positions}
        \label{fig:edge_positions_medmcqa}
    \end{subfigure}
    \hfill
    \begin{subfigure}[b]{0.32\linewidth}
        \includegraphics[width=\linewidth]{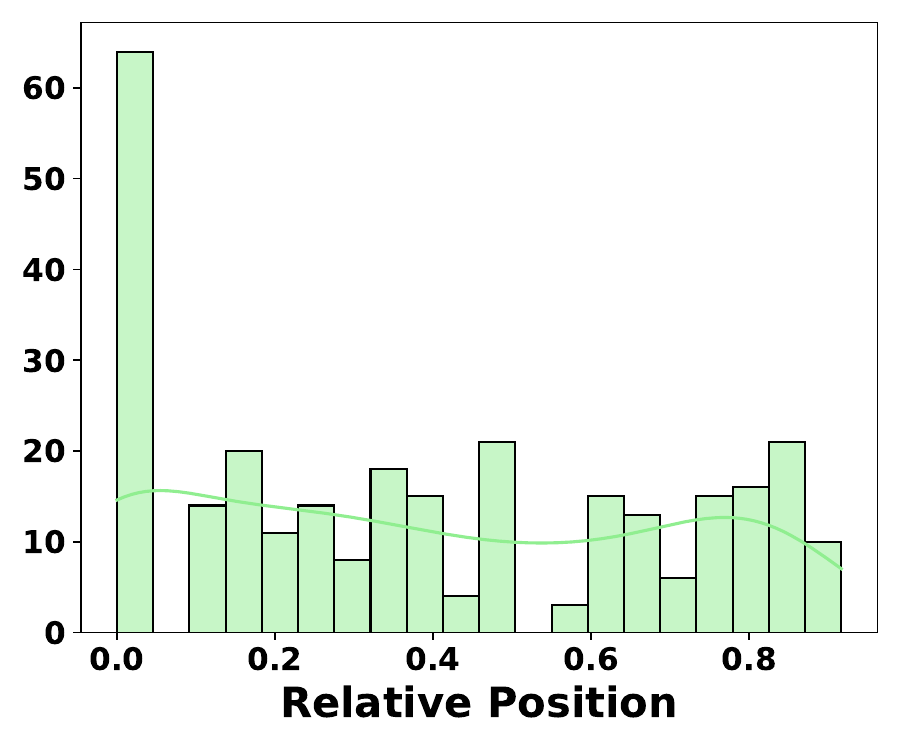}
        \caption{Subpath Positions}
        \label{fig:subpath_positions_medmcqa}
    \end{subfigure}

    \caption{Distribution of Critical Changes by Perturbation Position across datasets. Top: MMLU. Bottom: MedMCQA.}
    \label{fig:position_distribution_combined}
\end{figure}

Our experimental results, shown in Figure \ref{fig:position_distribution_combined}, suggest that perturbations occurring closer to the beginning of the path are more likely to cause significant changes in the generated output. This trend holds consistently across node, edge, and sub-path removals, in both datasets, indicating that early components in the path play a pivotal role in maintaining the stability of the solution. This trend is consistent with the design of our methodology, where the shortest path is computed from a designated start node to a fixed end node--both typically extracted from the user query and therefore semantically important. The influence of perturbations located in the middle or toward the end of the path is more variable, with no strong or consistent effect observed.

\captionsetup{labelfont=bf}
\begin{figure}[bh]
    \centering
    \begin{subfigure}[b]{0.45\linewidth}
        \centering
        \includegraphics[width=\linewidth]{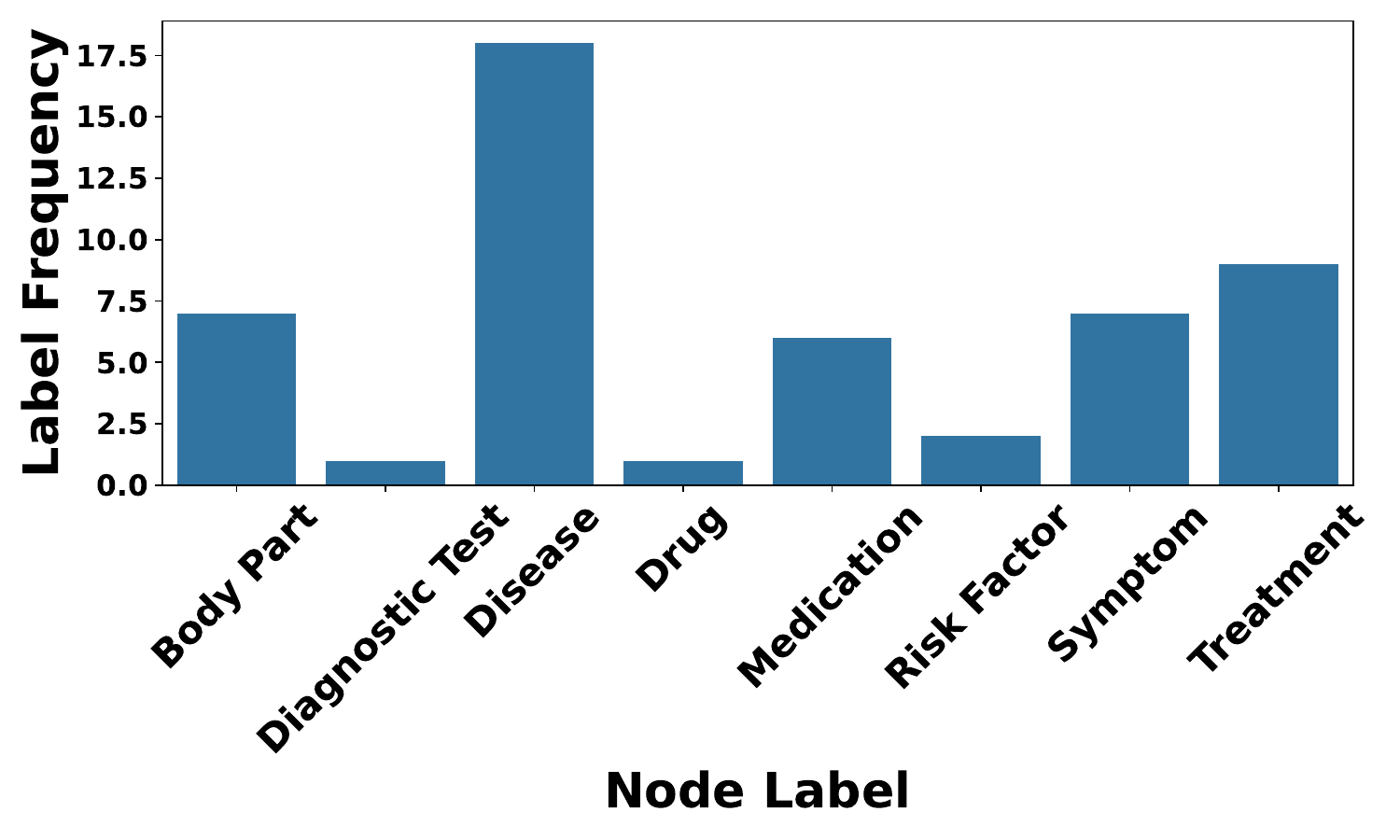}
        \caption{MMLU}
        \label{fig:label-dist-mmlu}
    \end{subfigure}
    \begin{subfigure}[b]{0.45\linewidth}
        \centering
        \includegraphics[width=\linewidth]{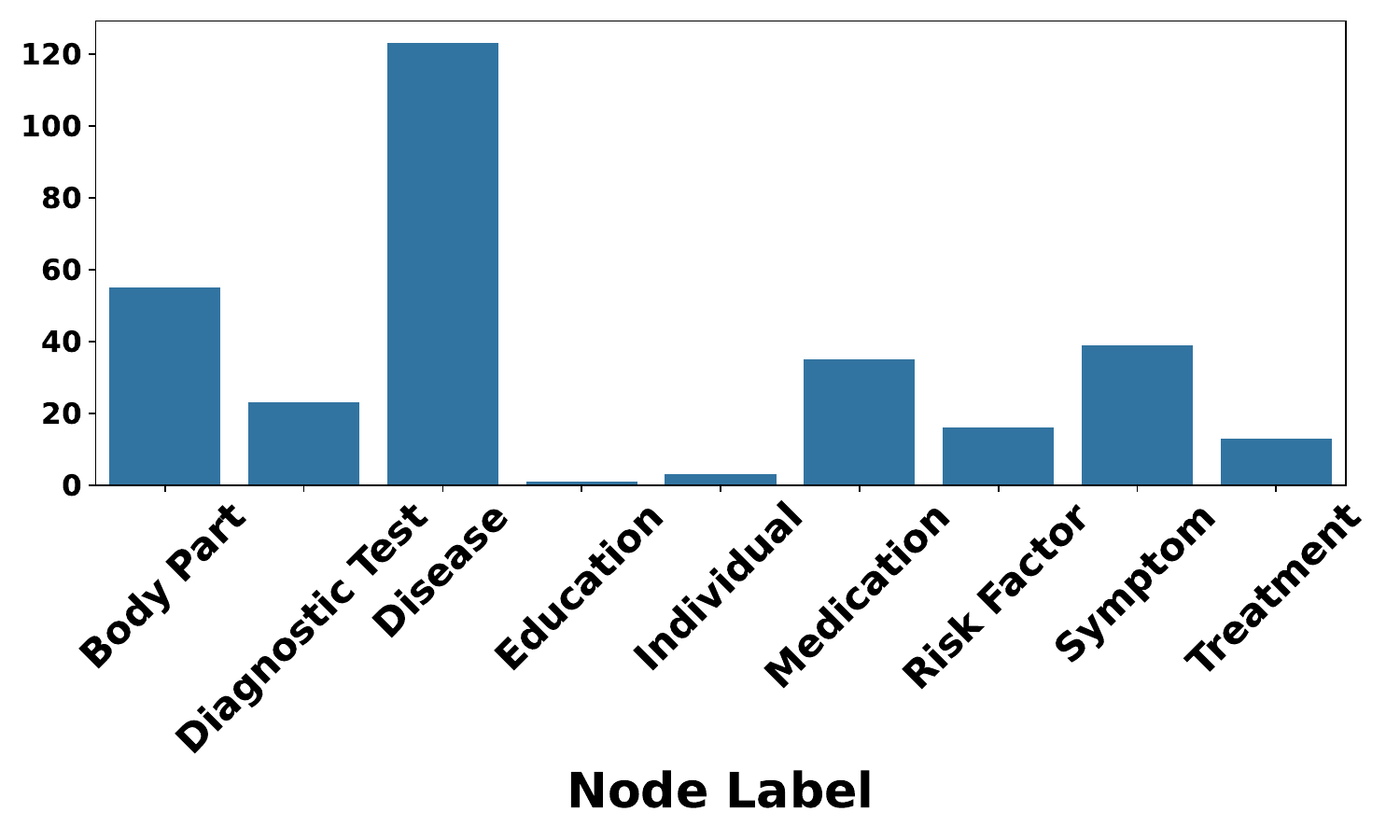}
        \caption{MedMCQA}
        \label{fig:label-dist-medmcqa}
    \end{subfigure}
    \caption{Distribution of critical changes by node labels across both datasets.}
    \label{fig:label-dist-both}
\end{figure}

\subsubsection{Assessing node significance via label analysis.}

Our knowledge graph includes labels for each node, such as Disease, Drug, Symptom, reflecting the semantic roles of different concepts within the graph. To understand the importance of these node types, we further analyzed how critical perturbations--those that cause output changes--are distributed across these labels. Specifically, we investigated whether certain categories of nodes are more likely to disrupt the output when removed.

We analyzed which node labels often triggered output changes. As shown in Figure \ref{fig:label-dist-both}, \textit{Disease}, \textit{Symptom}, and \textit{Body Part} consistently emerged as the most impactful across both MMLU and MedMCQA, highlighting their key role in model reasoning. More general or peripheral labels appeared less often, indicating a weaker influence on output stability.


By leveraging these node labels, our evaluation not only identifies structurally important components but also uncovers the semantic roles that drive the model behavior. This layer of interpretability allows us to better understand why certain perturbations matter--linking changes in output to specific domain-relevant concepts--thereby grounding our explanations in both graph structure and real-world meaning.


\begin{figure}[h]
    \centering
    \begin{subfigure}[b]{0.48\linewidth}
        \centering
        \includegraphics[width=\linewidth]{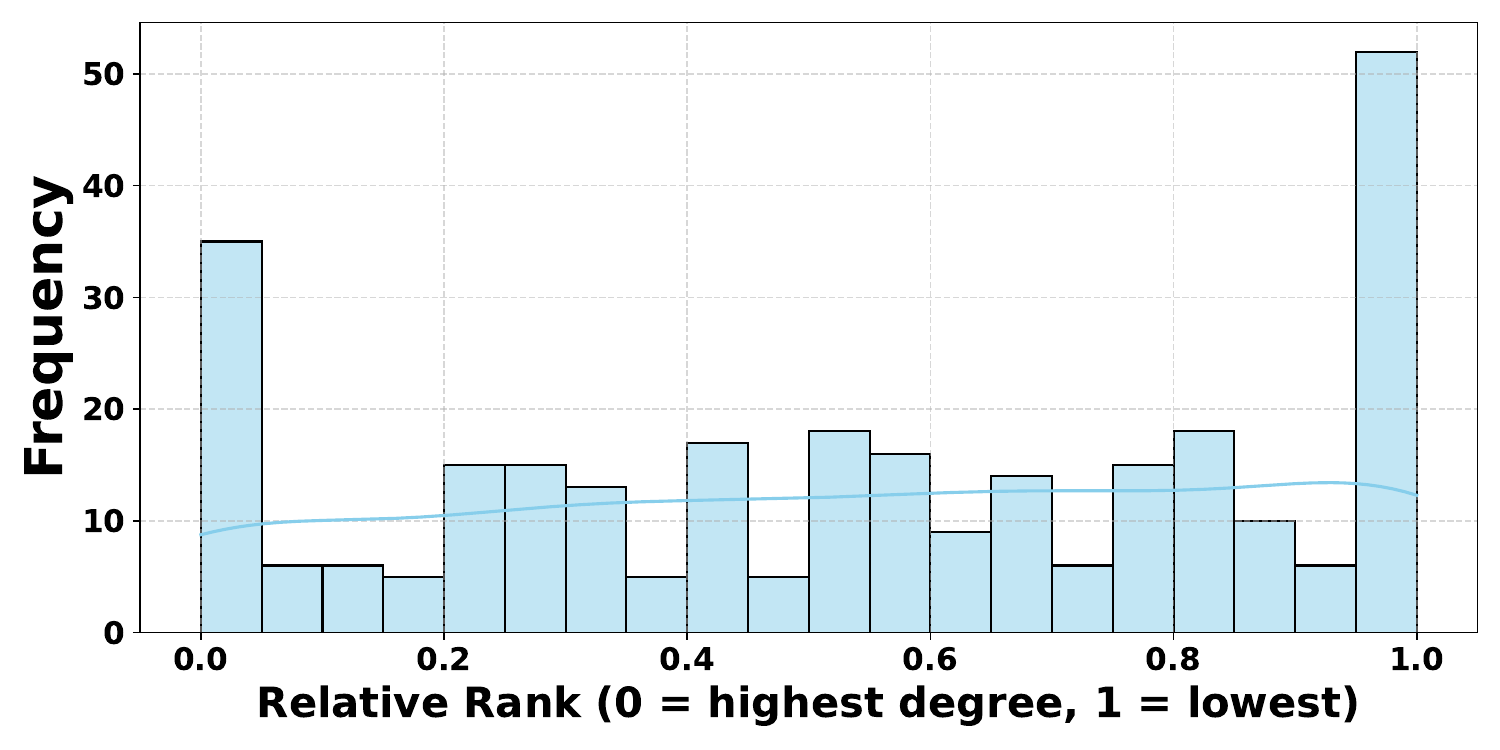}
        \caption{MedMCQA}
        \label{fig:node-rank-medmcqa}
    \end{subfigure}
    \hfill
    \begin{subfigure}[b]{0.48\linewidth}
        \centering
        \includegraphics[width=\linewidth]{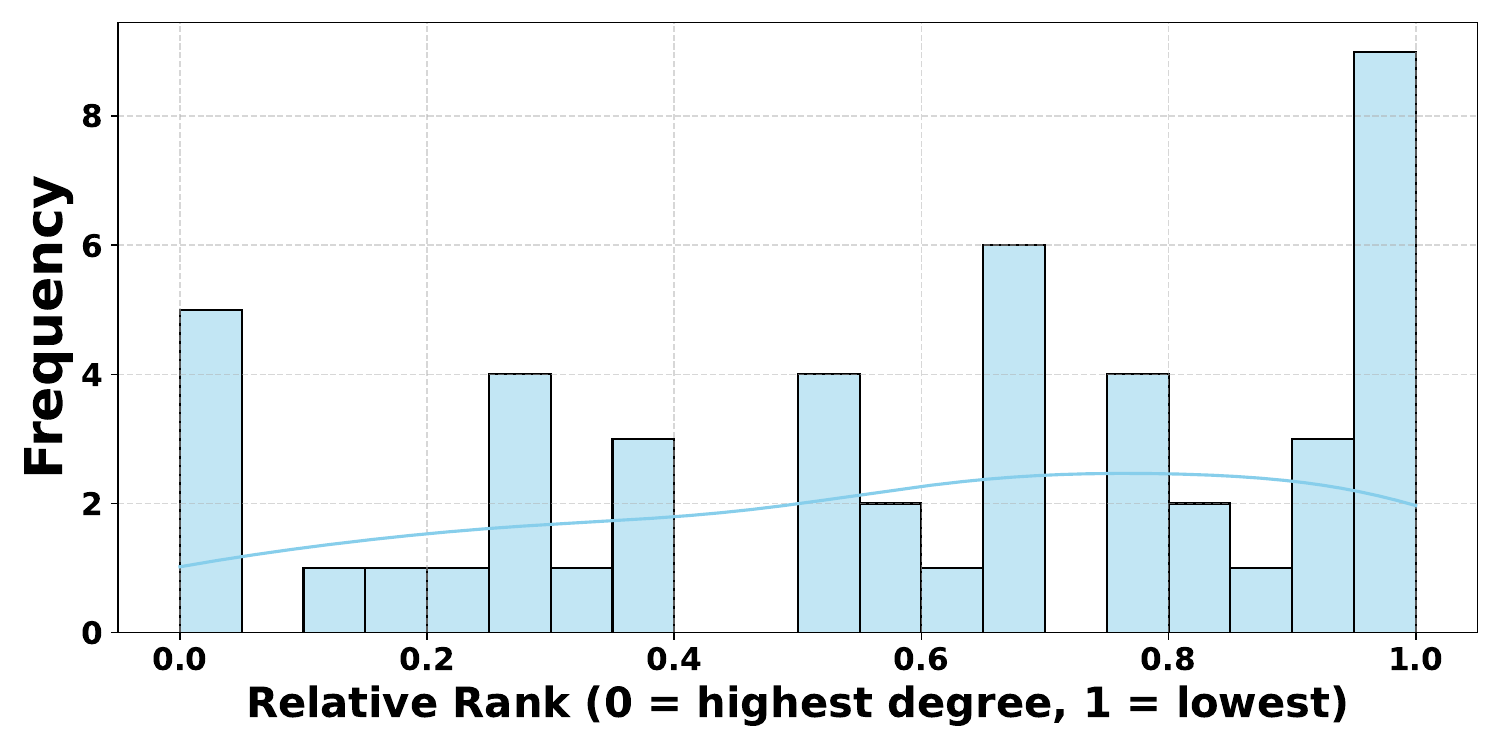}
        \caption{MMLU}
        \label{fig:node-rank-mmlu}
    \end{subfigure}
    \caption{Relative rank distribution of important nodes across two datasets, based on each node degree within the derived path.}
    \label{fig:important-node-rank-comparison}
\end{figure}

\begin{figure}[h]
    \centering
    \begin{subfigure}[b]{0.48\linewidth}
        \centering
        \includegraphics[width=\linewidth]{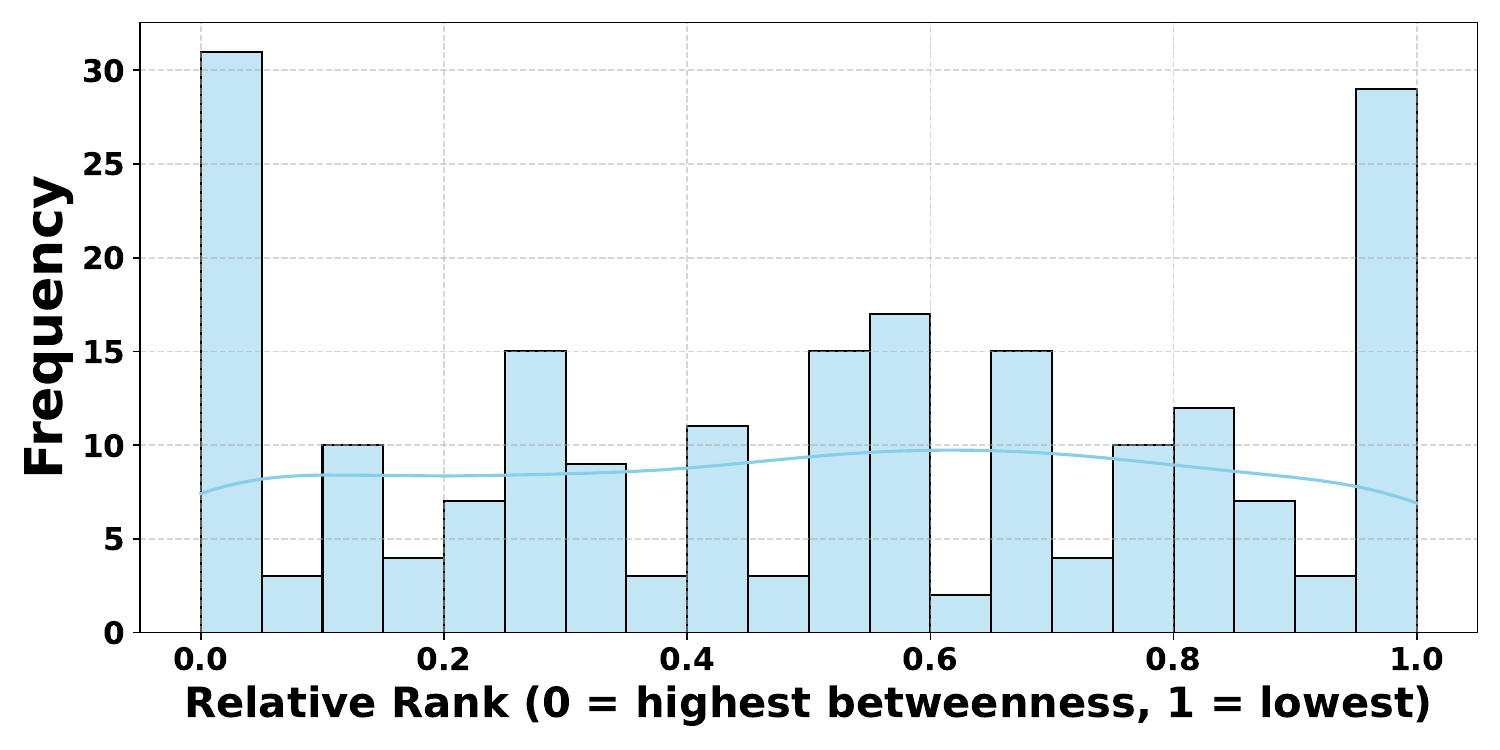}
        \caption{MedMCQA}
        \label{fig:edge-rank-medmcqa}
    \end{subfigure}
    \hfill
    \begin{subfigure}[b]{0.48\linewidth}
        \centering
        \includegraphics[width=\linewidth]{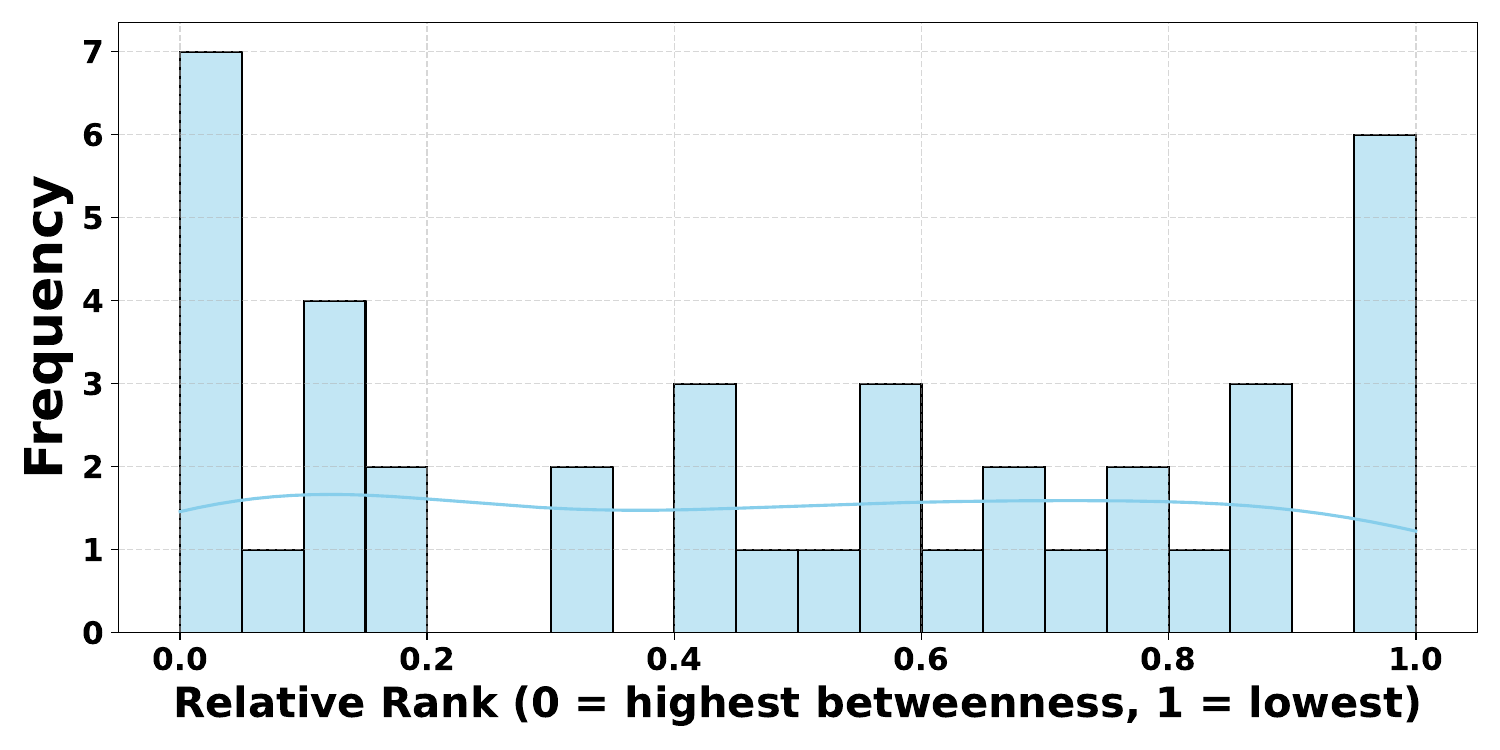}
        \caption{MMLU}
        \label{fig:edge-rank-mmlu}
    \end{subfigure}
    \caption{Relative rank distribution of important edges across two datasets, based on each edge betweenness score.}
    \label{fig:important-edge-rank-comparison}
\end{figure}

\begin{figure}[h]
    \centering
    \begin{subfigure}[b]{0.48\linewidth}
        \centering
        \includegraphics[width=\linewidth]{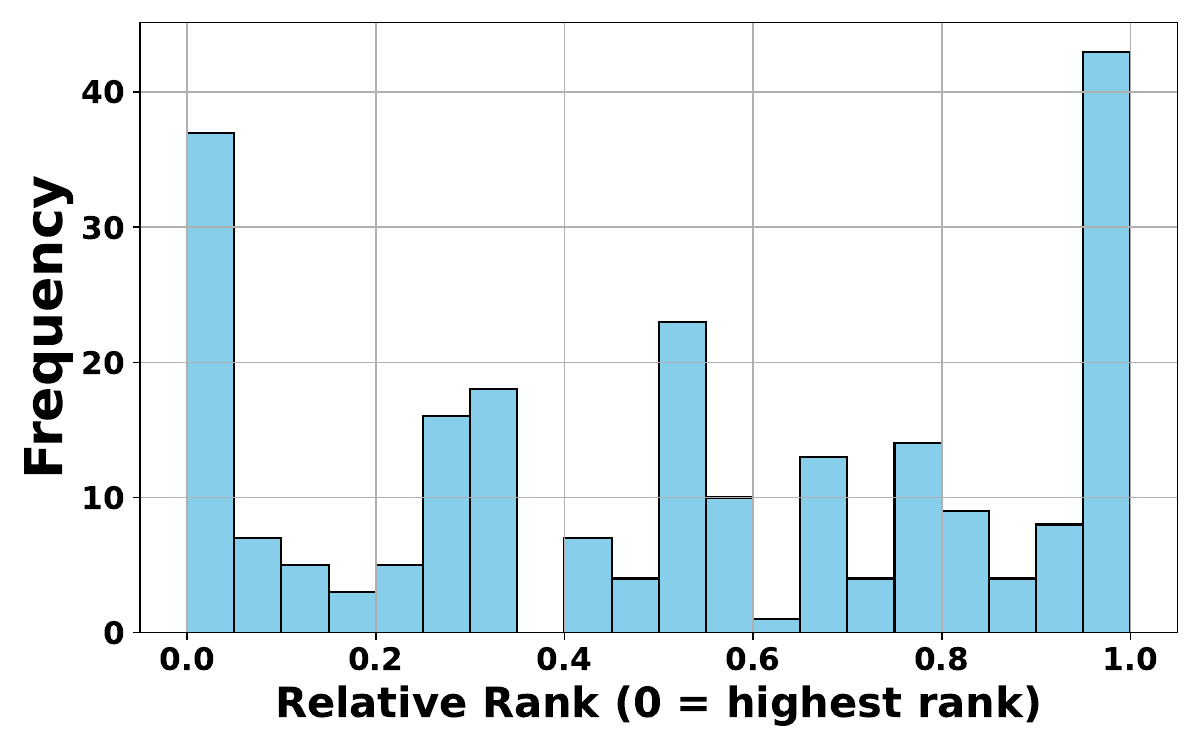}
        \caption{MedMCQA}
        \label{fig:subpath-rank-medmcqa}
    \end{subfigure}
    \hfill
    \begin{subfigure}[b]{0.48\linewidth}
        \centering
        \includegraphics[width=\linewidth]{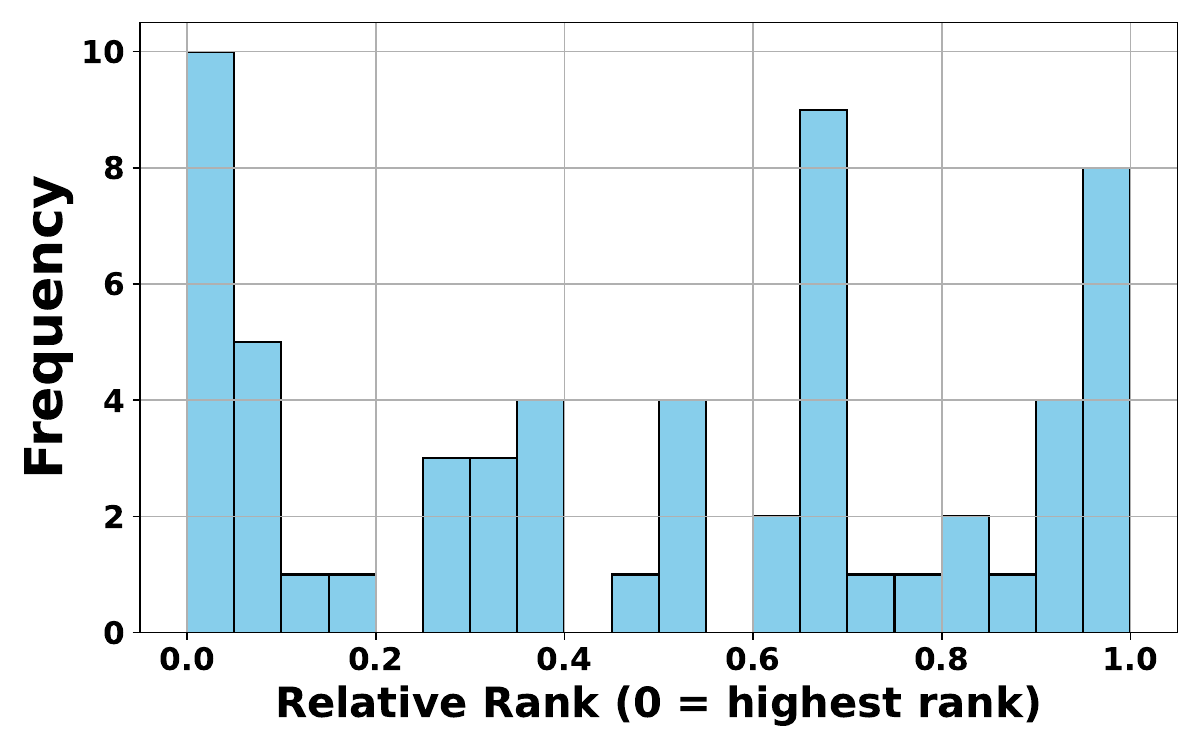}
        \caption{MMLU}
        \label{fig:subpath-rank-mmlu}
    \end{subfigure}
    \caption{Relative rank distribution of important sub-paths across two datasets, based on sub-path score.}
    \label{fig:important-subpath-rank-comparison}
\end{figure}

\subsubsection{Analyzing graph centrality metrics of important elements.} We now examine the relationship between the centrality of the elements in the KG and their importance in the query result.

\vspace*{0.1in}
\noindent \textit{Analyzing the degree of important nodes in the retrieved path.} The degree of a node can serve as a proxy for its importance within a knowledge graph. Highly connected nodes often function as central hubs whose removal may significantly alter the graph structure, while nodes with low degree can represent rare or specific entities that help differentiate context. In this analysis, we examine the degree of the identified important nodes within the derived shortest path to understand their structural roles and contextual relevance. 

To enable a normalized comparison between important nodes and all other nodes in the path, we ranked each node based on its degree, as shown in Figure \ref{fig:important-node-rank-comparison}. Across both datasets, nodes with the highest degree in the path (relative rank equal to 0) and those with the lowest degree (relative rank equal to 1) appear to impact the output response the most. As expected, central hubs and rare, context-specific nodes both play critical roles in guiding reasoning.

\vspace*{0.1in}
\noindent \textit{Examining the betweenness of important edges in the retrieved path.} While node degree captures local connectivity, edge betweenness centrality reflects the role of an edge in maintaining global graph flow. Edges with high betweenness are often critical conduits for information, lying on many of the shortest paths between pairs of nodes. In the context of our derived shortest paths, we examine the betweenness centrality of the identified important edges to assess their structural significance.

To compare their importance against others within the same path, we compute the relative ranks of each edge based on betweenness centrality. As shown in Figure \ref{fig:important-edge-rank-comparison}, the most impactful edges tend to occupy either extreme of the rank distribution--those with the highest betweenness (relative rank equal to 0) and those with the lowest (relative rank equal to 1). In this case, we observe a marginal increase in the concentration of important edges at rank 0 across both datasets, indicating a slight bias toward high-betweenness edges. This suggests that central edges--those frequently traversed in shortest paths--may have greater influence on the output, whereas low-betweenness edges may still play a complementary role by capturing niche or context-dependent relationships.

\vspace*{0.1in}
\noindent \textit{Evaluating the structural characteristics of important sub-paths.} Once important sub-paths have been identified, we further characterize their structural properties using a custom scoring function. Each sub-path is defined as a triplet consisting of two nodes and the connecting edge. For each important sub-path, we compute a score defined as the ratio of the edge betweenness centrality to the sum of the degrees of the two incident nodes:
\vspace{-0.2cm}
\begin{equation}
\text{Subpath Score} = \frac{\text{Betweenness of the Edge}}{\text{Degree of Node}_1 + \text{Degree of Node}_2}
\label{eq:score}
\end{equation}

By normalizing edge betweenness by the degrees of the connected nodes, this score balances global edge centrality with local node specificity. It serves as a useful heuristic to highlight edges that play key structural roles while linking less connected, and thus more contextually unique, nodes.

We rank all sub-paths within each retrieved path based on Eq. \ref{eq:score}, assigning relative ranks accordingly.
Figure \ref{fig:important-subpath-rank-comparison} illustrates the distribution of these ranks, showing that sub-paths positioned at the extremes tend to exert the greatest influence on the output of the model. 
This is consistent with the patterns seen for nodes and edges.  Overall, sub-paths with a high edge betweenness but low combined node degrees represent structurally central connections linking contextually unique nodes, while those with low edge betweenness but high node degrees indicate densely connected, but less structurally critical, parts of the graph.







\subsection{Comparative Analysis of KGRAG-Ex and RAG-Ex}

Existing frameworks, such as RAG-Ex \cite{10.1145/3626772.3657660}, employ perturbation-based strategies to generate explanations within a RAG environment. The central idea involves perturbing the RAG context and observing how these changes influence the output of the system. However, these perturbations are applied exhaustively, without a targeted or guided strategy, which significantly increases the overall computational cost.

In our work, to address this limitation, we introduced graph-level perturbation strategies, where modifications are performed by removing nodes, edges, and sub-paths (i.e., triplets) from the shortest path we derive based on the user query from our knowledge graph, rather than perturbing individual words or sentences in the retrieved context. This allows us to operate at a higher semantic level, focusing on structural elements that are more likely to influence the output. 

In this experiment, for RAG-Ex, we use a window size of 5, removing 5 tokens per perturbation. Smaller windows (removing 1–2 tokens) often had little impact on output, while larger windows (removing whole chunks or paragraphs) sometimes caused inconsistent effects, complicating reliable explanation generation. As shown in Table \ref{tab:two_dataset_comparison}, using KGRAG-Ex results in a significant reduction in the number of LLM calls and tokens consumed, leading to lower computational overhead and reduced pricing costs.

\setlength{\tabcolsep}{6pt}
\renewcommand{\arraystretch}{1.1}
\begin{table}[t]
\centering
\caption{Average LLM calls and  tokens for RAG-Ex and KGRAG-Ex across two datasets, MMLU and MedMCQA.}
\vspace{1em}
\begin{tabular}{|l|c|c|c|c|}
\hline
\multirow{2}{*}{\textbf{Method}} & \multicolumn{2}{c|}{\textbf{MMLU}} & \multicolumn{2}{c|}{\textbf{MedMCQA}} \\
                                 & \textbf{LLM Calls} & \textbf{Tokens} & \textbf{LLM Calls} & \textbf{Tokens} \\
\hline
RAG-EX                & 65 & 4032 & 61 & 3661 \\
KGRAG-Ex         & 20 & 2112 & 19 & 2180.5 \\
\hline
\textbf{Difference}          & 45 & 1920 & 42 & 1480.5 \\
\hline
\end{tabular}
\label{tab:two_dataset_comparison}
\end{table}

\section{Conclusions}
\label{sec:conclusion}

In this work, we presented KGRAG-Ex, a retrieval-augmented generation system that leverages knowledge graphs to improve both factual grounding and interpretability. Through graph perturbation-based explanation methods, we demonstrated how different components of the knowledge graph contribute to the reasoning process of the model. Our experiments provide insights into system sensitivity, the role of semantic node types, and the relationship between graph structure and explanation significance. Future work will explore extending the approach to larger and more diverse datasets, and investigating alternative perturbation techniques, such as adding edges, changing the directions of existing edges, and introducing non-existing nodes. 




%
%
%
%


\bibliographystyle{plain}
\renewcommand{\refname}{References}   
\bibliography{references}  

\begin{thebibliography}{10}

\bibitem{alkhamissi2022review}
Amir Alkhamissi, Mahmoud Abdel-Aziz, Hamdy Mahgoub, Mohamed Hammad, et~al.
\newblock A review of large language models: Applications, challenges, and
  opportunities.
\newblock {\em arXiv preprint arXiv:2207.02329}, 2022.

\bibitem{belinkov2022probing}
Yonatan Belinkov.
\newblock Probing classifiers: Promises, shortcomings, and advances.
\newblock In {\em Proceedings of the 60th Annual Meeting of ACL (Volume 1: Long
  Papers)}, pages 2735--2754. ACL, 2022.

\bibitem{bockling2024walkretrieve}
Martin Böckling, Heiko Paulheim, and Andreea Iana.
\newblock Walk\&retrieve: Simple yet effective zero-shot retrieval-augmented
  generation via knowledge graph walks.
\newblock In {\em Information Retrieval's Role in RAG Systems (IR-RAG 2025) in
  conjunction with SIGIR}, 2024.

\bibitem{danilevsky2020survey}
Marina Danilevsky, Yannis Qian, Michael Aharon, Yannis Katsis, Ilia Kuznetsov,
  Pranav Sen, Anbang Singh, and Partha~Pratim Talukdar.
\newblock A survey of the state of explainable ai for natural language
  processing.
\newblock {\em arXiv preprint arXiv:2010.00711}, 2020.

\bibitem{fan2024surveyragmeetingllms}
Wenqi Fan, Yujuan Ding, Liangbo Ning, Shijie Wang, Hengyun Li, Dawei Yin,
  Tat-Seng Chua, and Qing Li.
\newblock A survey on rag meeting llms: Towards retrieval-augmented large
  language models.
\newblock In {\em Proceedings of the 30th ACM SIGKDD Conference}, pages
  6491--6501, 2024.

\bibitem{jiang2023activeretrievalaugmentedgeneration}
Zhengbao Jiang, Frank~F Xu, Luyu Gao, Zhiqing Sun, Qian Liu, Jane Dwivedi-Yu,
  Yiming Yang, Jamie Callan, and Graham Neubig.
\newblock Active retrieval augmented generation.
\newblock In {\em Proceedings of EMNLP 2023}, pages 7969--7992, 2023.

\bibitem{kamalloo-etal-2023-evaluating}
Ehsan Kamalloo, Nouha Dziri, Charles Clarke, and Davood Rafiei.
\newblock Evaluating open-domain question answering in the era of large
  language models.
\newblock In Anna Rogers, Jordan Boyd-Graber, and Naoaki Okazaki, editors, {\em
  Proceedings of the 61st Annual Meeting of ACL (Volume 1: Long Papers)}, pages
  5591--5606, Toronto, Canada, July 2023. ACL.

\bibitem{DBLP:journals/corr/abs-2005-11401}
Patrick Lewis, Ethan Perez, Aleksandra Piktus, Fabio Petroni, Vladimir
  Karpukhin, Naman Goyal, Heinrich K{\"u}ttler, Mike Lewis, Wen-tau Yih, Tim
  Rockt{\"a}schel, et~al.
\newblock Retrieval-augmented generation for knowledge-intensive nlp tasks.
\newblock {\em NeurIPS}, 33:9459--9474, 2020.

\bibitem{lewisNEURIPS2020}
Patrick Lewis, Ethan Perez, Aleksandra Piktus, Fabio Petroni, Vladimir
  Karpukhin, Naman Goyal, Heinrich K\"{u}ttler, Mike Lewis, Wen-tau Yih, Tim
  Rockt\"{a}schel, Sebastian Riedel, and Douwe Kiela.
\newblock Retrieval-augmented generation for knowledge-intensive nlp tasks.
\newblock In H.~Larochelle, M.~Ranzato, R.~Hadsell, M.F. Balcan, and H.~Lin,
  editors, {\em NeurIPS}, volume~33, pages 9459--9474. Curran Associates, Inc.,
  2020.

\bibitem{naveed2024comprehensiveoverviewlargelanguage}
Humza Naveed, Asad~Ullah Khan, Shi Qiu, Muhammad Saqib, Saeed Anwar, Muhammad
  Usman, Naveed Akhtar, Nick Barnes, and Ajmal Mian.
\newblock A comprehensive overview of large language models.
\newblock {\em ACM Transactions on Intelligent Systems and Technology}, 2023.

\bibitem{10598017}
Joel Rorseth, Parke Godfrey, Lukasz Golab, Divesh Srivastava, and Jaroslaw
  Szlichta.
\newblock Rage against the machine: Retrieval-augmented llm explanations.
\newblock In {\em 2024 IEEE 40th International Conference on Data Engineering
  (ICDE)}, 2024.

\bibitem{10.1145/3626772.3657660}
Viju Sudhi, Sinchana~Ramakanth Bhat, Max Rudat, and Roman Teucher.
\newblock Rag-ex: A generic framework for explaining retrieval augmented
  generation.
\newblock SIGIR '24. Association for Computing Machinery, 2024.

\bibitem{query2doc}
Liang Wang, Nan Yang, and Furu Wei.
\newblock Query2doc: Query expansion with large language models, 2023.
\newblock Available at: \url{https://arxiv.org/abs/2303.07678}.

\bibitem{xiong-etal-2024-benchmarking}
Guangzhi Xiong, Qiao Jin, Zhiyong Lu, and Aidong Zhang.
\newblock Benchmarking retrieval-augmented generation for medicine.
\newblock In Lun-Wei Ku, Andre Martins, and Vivek Srikumar, editors, {\em
  Findings of ACL 2024}, pages 6233--6251, Bangkok, Thailand and virtual
  meeting, August 2024. ACL.

\bibitem{zhangTACLsummarization}
Tianyi Zhang, Faisal Ladhak, Esin Durmus, Percy Liang, Kathleen McKeown, and
  Tatsunori~B. Hashimoto.
\newblock Benchmarking large language models for news summarization.
\newblock {\em Transactions of ACL}, 12:39--57, 01 2024.

\bibitem{zhao-etal-2020-knowledge-grounded}
Xueliang Zhao, Wei Wu, Can Xu, Chongyang Tao, Dongyan Zhao, and Rui Yan.
\newblock Knowledge-grounded dialogue generation with pre-trained language
  models.
\newblock In Bonnie Webber, Trevor Cohn, Yulan He, and Yang Liu, editors, {\em
  Proceedings of EMNLP 2020}, pages 3377--3390, Online, November 2020. ACL.

\bibitem{zhu2025knowledge}
Xiangrong Zhu, Yuexiang Xie, Yi~Liu, Yaliang Li, and Wei Hu.
\newblock Knowledge graph-guided retrieval augmented generation, 2025.
\newblock arXiv:2502.06864.

\end{thebibliography}


\end{document}